%% file: main.tex
\documentclass[runningheads]{llncs}
\usepackage{graphicx}

\usepackage{tikz}
\usepackage{comment}
\usepackage{amsmath,amssymb} 
\usepackage{color}
\usepackage{graphicx}
\usepackage{amsmath}
\usepackage{amssymb}
\usepackage{booktabs}
\usepackage{graphicx}
\usepackage{textcomp}
\usepackage{algorithm}
\usepackage{algorithmic}
\usepackage{booktabs}
\usepackage{array}
\usepackage{bbm}
\usepackage{xcolor}
\usepackage{soul}
\usepackage{multirow}
\usepackage{enumitem}
\usepackage{adjustbox}
\usepackage{multirow}
\usepackage{multicol}
\usepackage{colortbl}
\usepackage{listings}
\usepackage{float}

\usepackage[accsupp]{axessibility}  

\DeclareMathOperator*{\argmax}{arg\,max}
  
\definecolor{Gray}{gray}{0.85}
\definecolor{index}{gray}{0.45}

\begin{document}
\pagestyle{headings}
\mainmatter
\def\ECCVSubNumber{6314}  

\title{Domain Generalization Emerges from Dreaming} 

\authorrunning{H. Heo et al.}
\author{Hwan Heo\inst{1} \and Youngjin Oh\inst{1} \and Jaewon Lee\inst{1} \and  Hyunwoo J. Kim \inst{1}\thanks{corresponding author.}}
\institute{1} 
\institute{\tt\small \{gjghks950, dign50501, 2j1ejyu, hyunwoojkim\}@korea.ac.kr}

\maketitle

\begin{abstract}
    \input{Content/0_Abstract}
\end{abstract}
\input{Content/1_Introduction}
\input{Content/2_Related_Works}

\input{Content/3_Methods}
\input{Content/4_Experiments}
\input{Content/5_Conclusions}
\clearpage
%
%
\bibliographystyle{splncs04}
\bibliography{egbib}

\input{Content/6_Supplementary}
\end{document}

%% file: Content/0_Abstract.tex
Recent studies have proven that DNNs, unlike human vision, tend to exploit \textit{texture} information rather than \textit{shape}. 
Such texture bias is one of the factors for the poor generalization performance of DNNs.
We observe that the texture bias negatively affects not only in-domain generalization but also out-of-distribution generalization, \textit{i.e.}, Domain Generalization.
Motivated by the observation, we propose a new framework to reduce the texture bias of a model by a novel optimization based data augmentation, dubbed \textit{Stylized Dream}.
Our framework utilizes adaptive instance normalization (AdaIN) to augment the style of an original image yet preserve the content.
We then adopt a regularization loss to predict consistent outputs between Stylized Dream and original images, which encourages the model to learn shape-based representations.
Extensive experiments show that the proposed method achieves state-of-the-art performance in out-of-distribution settings on public benchmark datasets : \textbf{PACS}, \textbf{VLCS}, \textbf{OfficeHome},  \textbf{TerraIncognita}, and \textbf{DomainNet}. 

%% file: Content/1_Introduction.tex
\section{Introduction}
\label{sec:intro}
{Domain Generalization aims to generalize \textit{out-of-distribution}, \textit{i.e.}, predicting well on the unseen target distribution.
The problem has been actively studied for the last few decades ~\cite{Balaji18metareg,antonio18dsam,Huang20RSC,Kim21selfreg,Li19epifcr,Matsuura20mmld,Nam21sagnet,nuriel21padain,Dou19msaf,Seo20DSON,sun2016coral,wang20eisnet,wang2021l2d,zhao20er,zhou21mix}.
Most of these methods are based on empirical risk minimization in an out-of-distribution generalization. 
However, the empirical risk minimization has an assumption that training distributions and test (real) distributions are similar. 
The assumption may not hold in real-world scenarios.
Therefore, it is suboptimal to generalize out-of-distribution by minimizing only the empirical risk.
Since this learning strategy depends on the losses of the training set, DNNs easily fall into the bias of the observed samples and often differ from human visions.
In other words, DNNs tend to exploit such bias as a  shortcut~\cite{gerihos2020shortcut}, which is a decision rule that performs well on in-domain generalization but fails when testing on examples from unseen distributions.}

Texture bias is one of the well-known shortcuts of DNNs.
Several studies~\cite{gatys2015texture,brendel2019bagnet,gerihos2019sin} have proven that DNNs tend to learn texture-based representations more than shape-based representations.
Besides, it has been found that this preference for textures leads to poor generalization performance of the DNNs.
Recent studies~\cite{gerihos2019sin,hermann2020texture,kim20texture1,Mummadi2021texture2} have empirically demonstrated that the texture bias of DNNs induces poor generalization performance. 
They have shown that training neural networks with uninformative texture-based representations shows better performance and robustness on human-like vision tasks.

We observed that this tendency also holds in Domain Generalization.
To verify that texture bias induces poor generalization performance, 
we conducted simple experiments in the following settings: training a model on Photo and evaluating it on Art Painting (P $\rightarrow $ A), and vice versa (A $\rightarrow $ P). 
Interestingly, our experiments show that generalization performance of P $\rightarrow $ A (\textbf{63.78}\%) is significantly lower than A $\rightarrow $ P (\textbf{91.44}\%), see Table~\ref{tab:res_single}.
Unlike images in Photo with the same class have similar textures, images in Art Painting with the same class have different textures. Also, in Art Painting, images from different classes often have the same texture. 
In other words, the textures of Art Painting are less informative than photos. 
As a result, the model trained in Art Painting has a better domain generalization performance than the model trained on Photo.


Inspired by the observation, we propose a learning method that reduces the texture bias of the model by using a novel optimization-based data augmentation method called \textit{Stylized Dream}. 
Similar to DeepDream~\cite{Mordvintsev2015deepdream}, which maximizes the activation of an input image, \textit{Stylized Dream} updates the image along a gradient-ascent direction of the feature norm.
Since DeepDream changes both the content (shape) and style (texture) of an original input image, we add an adaptive instance normalization (AdaIN)~\cite{huang1027adain} to disentangle each other.

During the training phase, to reduce the texture bias of the model, we utilize  consistency training by adding a regularization loss between original images and \textit{Stylized Dreams}.
Our framework minimizes both a supervised loss of the original image and consistency loss between the original image and Stylized Dream, which encourages the model to learn shape-based representations.
Our framework does not need an additional style-transfer model or datasets. Also, our framework trains a learner model in an end-to-end manner. 
\newline

\noindent Our \textbf{contributions} are threefold:
\begin{itemize}
    \item[\textbullet] We propose a novel optimization-based data augmentation method  \textit{Stylized Dream}, which augments the style of images while preserving their content. \vspace{3pt}
    \item[\textbullet] We show that the consistency regularization between \textit{Stylized Dream} and the original image reduces the texture bias of a model and improves domain generalization performance. \vspace{3pt}
    \item[\textbullet] Lastly, our experiments demonstrate that the proposed method achieves state-of-the-art performance on the public domain generalization datasets.
\end{itemize}

%% file: Content/2_Related_Works.tex
\section{Related Works}
\subsection{Domain Generalization}
The goal of domain generalization methods is to generalize well on the unseen target domain while training only with the source domains. 
A number of domain generalization methods~\cite{bahng2020learning,ganin2017dann,li2018cdann,muandet2013domain,sun2016coral,zhao20er} have tried to bridge the gaps between different domains in the feature space by matching the distributions of the domains. 
Meta-learning frameworks~\cite{Balaji18metareg,dou2019domain,li2018mldg} have learned the domain shift during training with pseudo-train and pseudo-test distribution. Kim et al.\cite{Kim21selfreg} has suggested self-supervised learning to generalize out-of-distribution by mapping images in the same class but different domains to the same representation space. 
SWAD~\cite{cha2021swad} shows that ERM training with stochastic weight averaging densely can achieve high domain generalizability to an unseen target domain.

Data augmentation is another way to improve the out-of-domain performance.
Shankar et al.~\cite{shankar2018generalizing} perturbs the input image by adding adversarial gradients deviated from a label and a domain classifier. DLOW~\cite{gong2019dlow} translates an image from one domain to an intermediate domain between the source domain and the target domain with a generative model. MixStyle~\cite{zhou21mix} mixes feature statistics of two different instances to generate new styles via convex combination.
Our method can be seen as one of the data augmentation strategies. 
Our work is similar to DLOW~\cite{gong2019dlow} in a way that both works translate an image to another domain. 
However, instead of using the generative model, our method perturbs an image with adversarial attack by leveraging adaptive instance normalization~\cite{huang1027adain}.

\subsection{Texture and Shape}
Geirhos et al.~\cite{gerihos2019sin} show that CNNs are biased towards textures rather than shapes. They reduce the texture bias of the DNNs by generating a dataset while the local textures are no longer an important cue for predicting the class label. CNN trained with this dataset improves the performance on object detection. Also, it shows robustness towards a wide range of image distortions. Brendel et al.~\cite{brendel2019bagnet} has supported this research with a new architecture called BagNets that is trained by restricting the receptive field to small local image patches, performing reasonably well on ImageNet. Hermann et al.~\cite{hermann2020texture} uses data augmentation such as color distortion, noise, and blur and remove random-crop augmentation to decrease texture bias. Unlike previous methods, our method uses optimization based data augmentation to augment the images in an end-to-end manner.

\subsection{Consistency training}
{Consistency training is one of approaches to semi-supervised learning.
Consistency training methods~\cite{athiwaratkun2018there,bachman2014learning,berthelot2019mixmatch,clark2018semi,laine2016temporal,luo2018smooth,miyato2016adversarial,miyato2018virtual,sajjadi2016regularization,tarvainen2017mean,verma2019interpolation,xie2020unsupervised} mainly constrain model predictions, regarded as a confidence indicator, to be invariant to small data augmentation of the input, or small noise applied to hidden states or model parameters.
Consistency training combine a supervised loss with an unsupervised consistency loss term (\textit{e.g.}, Mean Squared Error, KL divergence) that encourages the same output distribution in response to unsupervised samples augmented.

$\mathbf{\Pi}$~\cite{laine2016temporal} Model passes each sample to a model twice applying different augmentation and dropout, and then minimizes the mean square differences between the two resulting prediction vectors. 
Mean teacher method~\cite{tarvainen2017mean} proposed teacher network whose weights are in exponential moving average of the weights of the student and applies consistency cost between predictions of student and teacher model.
ICT~\cite{verma2019interpolation} and MixMatch~\cite{berthelot2019mixmatch} mix data points using Mixup~\cite{zhang2017mixup} to encourage consistency with the mixed predictions. 
UDA~\cite{xie2020unsupervised} utilizes rich data augmentation via RandAugment~\cite{cubuk2020randaugment} and back-translation~\cite{edunov2018understanding,sennrich2015improving}.
}

%% file: Content/3_Methods.tex
\section{Method}
\label{section:3}
{
In this section, we introduce our method that improves the domain generalizability of the neural network by leveraging a novel data augmentation framework.
Before introducing the main method, we first briefly describe domain generalization.
Figure~\ref{fig:main} illustrates an overview of the proposed method.
}

\input{3_Methods/3_1_Domain_Generalization}
\input{3_Methods/3_3_Dreaming}
\input{3_Methods/3_4_Adv_Training}

%% file: 3_Methods/3_1_Domain_Generalization.tex
\subsection{Domain Generalization}
\label{section:3.1.1}
\input{Figures/main_figure}
Consider an image $x$ in multiple source domains $\{  X_s \}_{s \neq t}$ where $t$ is the target domain and its label $y \in  Y$ for total $C$ classes. 
The goal of domain generalization is to learn domain-invariant representations that lead to more robust predictions on a target domain $ X_t$ as well.
In the out-of-distribution setting, the deep neural network with its parameter  $\theta$ is trained via empirical risk minimization as,
\begin{equation}
    \label{eq:DG}
    \begin{split}
        &\min_\theta \mathbb E_{x \sim \{  X_s \} _{s \neq t}, \ y \sim  Y} \ \  \mathcal L( x,\ y \ ; \theta),  \\
    \end{split}
\end{equation}
where $\mathcal L$ denotes a surrogate loss function (\textit{e.g.}, cross-entropy loss)  for the optimal classification loss. 
Since the target domain $ X_t$ is inaccessible during training, the majority of the previous works assume the existence of statistical invariance across source and target domains, and leverage the property to improve the generalization performance of the models.

%% file: Figures/main_figure.tex
\begin{figure}[t]
    \centering
    \includegraphics[width=\columnwidth]{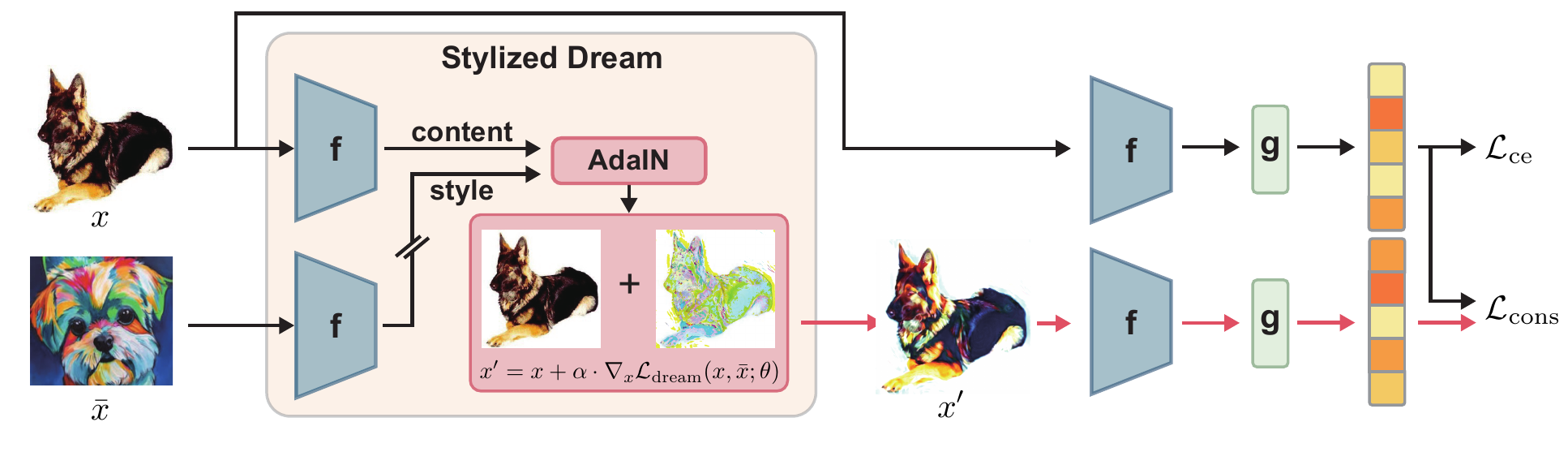}
    \caption{\textbf{Overview of our proposed method.} 
    Given an input image $x$, we randomly sample an image $\bar x$ with the same label. 
    Then these images are fed to the $f$ to extract feature maps $z$ and $\bar z$.
    $z$ is aligned with the channel-wise mean and standard deviation of $\bar z$ (Equation~\eqref{eq:dream}).
    Then we generate a \textit{Stylized Dream} image via maximizing the norm of the aligned $z$. 
    Lastly, clean image $x$ and \textit{Stylized Dream} image $x^\prime$ are fed into the networks $f$ and $g$ to obtain logits. 
    The logit of clean image $x$ is utilized to get training loss $\mathcal L_\text{ce}$. 
    The consistency loss $\mathcal L_\text{cons}$ takes the logit of clean image $x$ and that of \textit{Stylized Dream} image $x^\prime$.
    }
    \label{fig:main}
\end{figure}

%% file: 3_Methods/3_3_Dreaming.tex
\subsection{Stylized Dream}
\label{sec:SDA}
As aforementioned in Section~\ref{sec:intro}, DNNs have a strong bias towards the style of an image (\textit{i.e.,} texture) rather than the content of an image (\textit{i.e.,} shape).
We aim to reduce the texture bias of a network to achieve better domain generalization performance.
To this end, our goal is to augment the style of the input image while preserving its content. 
Since explicitly disentangling styles and contents is intractable, we utilize statistics of the features as style representation,
which is proven to be effective for style transfer~\cite{huang1027adain}.
\input{Algorithms/Alg1}

Consider a classification network with a feature extractor $f$ and a linear classifier $g$. 
Let $z \in \mathbb R ^{ D \times H \times W}$ be the feature maps of image $x$ extracted with the feature extractor $f$, \textit{i.e.}, $z = f(x)$ where $H$ and $W$ denote spatial dimensions of the feature maps and $D$ is the number of channels.  
Then, the channel-wise mean and standard deviation of $z$, which are the statistics of the feature maps, are derived by,
\begin{equation}
    \label{eq:AdaIN}
    \begin{split}
    \mu_z^c = {\frac{1}{ HW}} \sum_{h=1}^{ H} \sum_{w=1}^{ W} z^{chw}, \;\;\;\;
        \sigma_z^c = \sqrt{ {\frac{1}{HW}} \sum_{h=1}^{ H} \sum_{w=1}^{ W} (z^{chw} - \mu_z^c ) ^2 }.
    \end{split}
\end{equation}
Recent works in style transfer use these statistics to replace instance-wise styles while preserving their contents. 
Based on the intuition, we define a loss function for generating style-augmented examples as follows: 
\begin{equation}
    \label{eq:dream}
    \begin{split}
        \mathcal{L}_\text{dream}(x, \ \bar x \ ;\theta) = \left\|   (f(x) - \mu_{f(x)} ) \frac{\sigma_{f(\bar x)}}{\sigma_{f(x)}} +  \mu_{f(\bar x)}  \right\|_F ,
    \end{split}
\end{equation}
where $x$ is the content input image, $\bar x$ is an arbitrary style input image that has the same label as $x$. 
To avoid the numerical instability of division with a close-to-zero number in AdaIN, we add a small value $\epsilon=10^{-5}$ to the denominator as in the original AdaIN implementation. 

In Equation~\eqref{eq:dream}, the mean and variance of the feature map $z$ are aligned to match those of the feature maps $\bar z$~\cite{huang1027adain}, \textit{i.e.}, adaptive instance normalization (AdaIN).
Since we sample $\bar x$ to have the same label as $x$, its content (label) is preserved. 
Then, finding the \textit{Stylized Dream} (SD) $x'$ can be formulated as an optimization problem as below,
\begin{equation}
    \label{eq:SD_objective}
    \begin{split}
        x' &= \  x + \delta',~~\\ 
         \text{s.t. } \delta' & = \argmax_\delta \ \mathcal{L}_\text{dream}(x + \delta, \ \bar x \ ;\theta). \\ 
    \end{split}
\end{equation}
We solve this optimization problem via the projected gradient ascent as follows:
\begin{equation}
    \label{eq:dream_attack}
    \begin{split}
        x^{i+1} &=  {\rm \Pi}_{\Delta} \left ( x^i + \alpha \cdot  \nabla _{x^i} \mathcal L_{\text{dream}} (x^i,\ \bar x\ ;\theta) \right ), 
    \end{split}
\end{equation}
where $x^0 = x$ and $\Pi_{\Delta}$ denotes the projection to valid RGB space $\Delta$ which ranges from 0 to 255. 
After $K$ iterations, SD example $x'=x^K$ is obtained.

\textit{Deep Dream}~\cite{Mordvintsev2015deepdream} directly maximizes the norm $\|f(x) \|_F$ instead of Equation~\eqref{eq:dream}, which enhances the activation of features in the certain layer of an image. 
This strengthens the original feature regardless of style and content.
In contrast to the \textit{Deep Dream}, maximization of the adaptive normalized feature norm weakens its original style yet enhances the original content. 
Algorithm~\ref{alg:attack} depicts how the \textit{Stylized Dream} (SD)  works.
We visualize our Stylized Dream examples (See Section~\ref{subsubsec:qual}).
Also, we demonstrate that our method is more effective than \textit{Deep Dream} for domain generalization (See Section~\ref{subsec:ablation}). 

%% file: Algorithms/Alg1.tex

\begin{algorithm}[t]
    \caption{\textbf{Stylized Dream}}
    \label{alg:attack}
    \textbf{Input:} {feature extractor $f$ parameterized by $\theta$, content image $x$, style image $\bar x$, step size $\alpha$ and iteration $K$.${}$}
    \begin{algorithmic}[1]
        \FUNCTION{{StylizedDream}$(f, x,\bar x, \alpha, K)$}
        \STATE $x' \leftarrow x$
        \FOR {$i$ in $1,\dots,K$}
            \STATE  $x' \leftarrow {\rm \Pi}_{\Delta} \left ( x' +  \alpha \nabla_{x'} \mathcal{L}_\text{dream} (x', \ \bar x \ ;\theta) \right)$
        \ENDFOR
        \STATE return $x'$
        \ENDFUNCTION
    \end{algorithmic}
\end{algorithm}

%% file: 3_Methods/3_4_Adv_Training.tex
\subsection{Training with Stylized Dream}
\label{sec:SDA_train}
We now present a learning strategy to train a network to predict consistent output between original images and \textit{Stylized Dream}.
Domain generalization aims to learn conditional distribution $P(Y|f(X))$ that is invariant to changes of the marginal distribution $P(X)$ across sources and target domains.
As aforementioned in Section~\ref{sec:intro}, we experimentally demonstrate that the texture bias induces poor generalization performance, \textit{i.e.,} the less texture biased model generalizes better. 
Motivated by the observation, we aim to make the feature extractor $f$ in $P(Y|f(X))$ to be independent to the textures.

Consequently, our goal is to make the output distribution of \textit{Stylized Dream} $p_\theta(y \ | \ x')$ consistent with the output distribution of source domains $p_\theta(y \ | \ x)$ where $p_\theta(y \ | \  x) = g(f(x))$.
Since \textit{Stylized Dream} and original images have the same content but different textures, consistent prediction between \textit{Stylized Dream} and original image encourages the model to learn texture invariant features. 
For the purpose, the objective for training the network to be less texture biased can be formulated as follows: 
\begin{equation}
    \label{eq:consistency_reg}
    \begin{split}
        \min_{\theta} \ & \mathbb E_{x \sim \{  X_s \} _{s \neq t}, \ y \sim  Y} \big [\ D \left( p_\theta(y \ | \ x') ,\    p_\theta(y \ |\ x ) \right)\ \big] \\
        \text{  s.t. } & x' = x + \delta^* \\
        & \delta ^{*} = 
        \argmax_{\delta} \mathcal{L}_\text{dream} (x+\delta,\ \bar{x} \ ;f) ,
    \end{split}
\end{equation}
where $D$ denotes the divergence metric between distributions, $\bar{x}\neq x$ is an arbitrary image with the same label as $x$.
Equation~\eqref{eq:consistency_reg} makes the model less texture biased by matching the original images with the augmentations of uninformative texture.

Combining with a standard task-related supervised loss, our loss function can be rewritten as,
\begin{equation}
    \label{eq:training_loss}
    \begin{split}
        & \min_{f,\ g}  \ 
        \mathbb{E}_{x \sim \{  X_s \} _{s \neq t}, \ y \sim  Y} \left[ \  \mathcal{L}_{\text{ce}}(x,\ y \ ; f,g) 
        + D
        \big{(} g(f(x')) 
        ,\ g(f(x)) \big{)}  \ 
        \right] ,
    \end{split}
\end{equation}
where $\mathcal L_{\text{ce}}$ denotes the cross-entropy loss for supervised learning.
Here, $\mathcal L_{\text{ce}}$ encourages the model to learn general representations from multiple source domains.

Similar to knowledge distillation, we use smoothed logits of the prediction (soft label) in training. Our final training loss function is written as,
\begin{equation}
    \label{eq:trades}
    \begin{split}
        &\min_\theta\mathbb E_{x, y} \bigg\{ \mathcal L_{\text{ce}} ( x,\ y \ ; \theta) +  \mathcal L_{\text{cons}} \big(\hat y^\prime / \tau,\ \hat y / \tau   \big)  
        \bigg\} ,
    \end{split}
\end{equation}
where $\mathcal L_{\text{cons}}$ is a consistency loss as a divergence metric, we use MSE loss, JS divergence, and KL divergence loss in experiments. 
$\hat y ^\prime$ and $\hat y $ denote a predicted logit of the \textit{Stylized Dream}  and original image, and $\tau$ denotes  temperature parameter. 
\input{Algorithms/Alg2}
Algorithm~\ref{alg:training} shows our training scheme with SD. \\

{
\noindent{\textit{\textbf{Remarks.}}}
Our framework is relevant to standard adversarial training in the sense that both methods generate augmented samples by numerical optimization schemes. 
However, adversarial training is different from our method in the two aspects. 
First, adversarial training generates the worst-case examples by maximizing the original supervised loss whereas our Stylized Dream (SD) 
generates an augmented sample by maximizing the activation of a stylized image, which is independent of the original loss.
Second, our framework encourages the consistency between predictions on clean and augmented samples by consistency regularization whereas the adversarial training directly optimizes the original supervised loss with the augmented samples.
}

%% file: Algorithms/Alg2.tex
\begin{algorithm}[t]
\caption{\textbf{Training with Stylized Dream}}
\label{alg:training}
\textbf{Input} {Training epochs $N$, Batch size $B$, learning rate $\eta$, smooth temperature $\tau$, step step $\alpha$, iteration $K$,  feature extractor $f$ and classifier $g$ parameterized by $\theta_f$ and $\theta_g$ respectively. }
\begin{algorithmic}[1]
    \STATE $\theta = \{ \theta_f, \ \theta_g \}$
    \FOR{$t=1,\dots,N$}
        \FOR{$i=1,\dots,B $ (in parallel)}
            \STATE get $i$th input image $x_i$ and its label $y_i$
            \STATE $\bar x_i \leftarrow$ randomly sampled image with label $\bar y_i = y_i$ 
            \STATE $x_i^\prime \leftarrow \text{StylizedDream}(f,x_i,\bar x_i,\alpha,K)$
            \STATE $\hat y_i \leftarrow g(f(x_i))$
            \STATE $\hat y_i^\prime \leftarrow g(f(x_i^\prime))$
        \ENDFOR
        \STATE $\theta \leftarrow \theta -
        {\frac{\eta}{B}} \! \cdot \! \sum^B_{i=1} \!\! \nabla_\theta \!\! \ \left[ \
        \mathcal L_\text{ce}(x_i,\ y_i\ ;\theta) \!+ \!
         \mathcal L_\text{cons}(\hat y_i^\prime/\tau,\ \hat{y_i}/\tau) \ \right] \!$
    \ENDFOR
\end{algorithmic}
\end{algorithm}

%% file: Content/4_Experiments.tex
\section{Experiments}
In this section, we empirically evaluate the effectiveness of our framework with \textit{Stylized Dream}.
We first briefly introduce datasets and provide implementation details. 
Before the main results of domain generalization, we provide a visualization of our Stylized Dream.
In the main results, we first report the experimental results of single-source domain generalization performance to verify our assumption: 'The less texture-biased model generalized better'.
Then, we demonstrate that the model trained with Stylized Dream considerably outperforms the previous state-of-the-art approaches in various public domain generalization benchmark datasets. 
Also, we present ablation studies of our work:  (1) Ablation studies to examine the role of each element in our method, (2) Ablation studies about divergence metrics and baseline model selections.
\input{4_Experiments/4_1_Implementations_Details}
\input{4_Experiments/4_2_Main_Results}
\input{4_Experiments/4_3_Ablations}

%% file: 4_Experiments/4_1_Implementations_Details.tex
 \subsection{Implementation Detail}
\subsubsection{Datasets and Baseline Models.}
Following Domainbed~\cite{Gulrajani21domainbed}, 
we evaluate and compare our method on various benchmarks:
\textbf{PACS}~\cite{li2017pacs}: 9991 images with 7 classes and 4 domains, \textbf{VLCS}~\cite{fang2013vlcs}: 10729 images with 5 classes and 4 domains, and \textbf{TerraIncognita}~\cite{beery2018terra}: 24,788 images with 10 classes and 4 domains.
Also, we report results on public domain adaptation datasets with out-of-distribution setting:  \textbf{Officehome}~\cite{venkateswara2017officehome}: 15,588 images with 65 classes and 4 domains, and  \textbf{DomainNet}~\cite{peng2019domainnet}: 586,575 images with 345 classes and 6 domains.

For the fair comparison with prior works, we train ResNet-50~\cite{he2015resnet} for our baseline model.
In training, the feature extractor and batch normalization are pre-trained on ImageNet~\cite{Deng2009imagenet} while the linear classifier is fully trained on the domain generalization dataset.
Unlike many other previous works~\cite{Seo20DSON,Kim21selfreg,Gulrajani21domainbed}, we do not freeze the batch normalization layers.
Following SelfReg~\cite{Kim21selfreg} and SagNet~\cite{Nam21sagnet}, we train our network for 30 epochs by SGD optimizer, with the learning rate set to 0.004, weight decay to 1e-4, momentum to 0.9, and training batch to 128.
In \textit{Stylized Dream}, step size $\alpha$ is set to 0.3 and iteration $N$ to 1. 
Since we perform SD with a single iteration, it demands only one extra backward step, so the increase of resource consumption from the benign training is marginal.  

\subsubsection{Evaluation on Out-of-distribution.}
Following the Gulrajani et al.~\cite{Gulrajani21domainbed}, 
we evaluate domain generalization performance in out-of-distribution setting for each domain and report the average performance. 
Each model is only trained on the multiple source domains and tested on the target domain. 
We average the accuracy scores of three independent runs for each domain, selecting each model with the lowest validation loss.
The rest of the numbers are from the original literature and re-implemented scores in the Domainbed~\cite{Gulrajani21domainbed}.

%% file: 4_Experiments/4_2_Main_Results.tex
\subsection{Qualitative Results.} 
\label{subsubsec:qual}
\noindent{\textbf{Visualization of Stylized Dream.}} 
\input{Figures/SD}
We provide visualization of our \textit{Stylized Dream} across the different domains and classes in the PACS dataset.
The visualization of our model is shown in Fig.~\ref{fig:SD}.
The figure is divided into three rows denoting the original image, arbitrary style image, and its \textit{Stylized Dream} respectively. 
All the Stylized Dream examples are visualized using the VGGNet-16 feature extractor with 10 iterations and 0.09 of step size $\alpha$. 
In Fig.~\ref{fig:SD}, our \textit{Stylized Dream} effectively augments its original texture while preserving its shape.
As discussed in the literature~\cite{Gatys16transfer}, the features of deep residual networks are less effective for style transfer or stylization. \\

\noindent{\textbf{Shape-based representation of Stylized Dream.}} 
{As argued by Li \textit{et al.}~\cite{Yingwei2020homogeneity}, if the neural networks focus more on the \textit{shape-based} representations, the adversarial perturbations of those models reveal a coarser level of granularity. 
\input{Figures/shape.tex}
The perturbations are more locally correlated and more structured than the naturally trained model (\textit{texture-biased}).
In other words, we can indirectly examine whether the model learns shape-based representations via adversarial perturbation on the model.
To investigate a shape-based representation of our method, we have conducted an adversarial attack on the baseline model (vanilla ResNet) and the proposed method.
As shown in Figure~\ref{fig:adv}, adversarial perturbations (by PGD) reveal that our model better recognizes the contour of the foreground than the naturally trained model. 
The proposed method encourages the network to focus on shape rather than texture and thus enhances out-of-distribution generalization. }

\subsection{Main Results}
\label{sec:main_results}
\subsubsection{Single-source Domain Generalization.}
As discussed in Section~\ref{sec:intro}, we conduct an experiment with the domain generalization performance of the model learned with a single-source domain in the PACS dataset.
We report scores for each one-to-one source-target combination with the ResNet-18 architecture.
In Table~\ref{tab:res_single}, each row and column indicates the training source and target domain.
As shown in Table~\ref{tab:res_single}, the performance difference between the Photo to the Art Painting (63.87\%) and the Art Painting to the Photo (91.44\%) implies that the texture bias negatively affects the domain generalization.
Moreover, the Photo-only trained model's average accuracy (40.2\%) shows the worst domain generalization performance across the domain.
It is even worse than the Sketch domain, which has few visual cues (only edges). 
This clearly indicates that the model trained with uninformative textures shows better  generalization performance.
\input{Tables/8_res_single}
\input{Tables/1_Single}

Moreover, we report the single-source domain generalization performance of our method.
As for the baseline, we compare the performance with SelfReg~\cite{Kim21selfreg} on the same setting.
As shown in Table~\ref{tab:single}, our experimental results are also effective in the extreme cases of domain generalization, outperforming prior works by 3.97\% on average when trained on ResNet-18.

\subsubsection{Performances on Various Benchmark Datasets.}
\input{Tables/0_Main}

In Table~\ref{tab:pacs}, we report the out-of-domain accuracies on PACS datasets compared with the previous state-of-the-art methods. 
We report the full table based on their baseline model of ResNet-50.
As shown in Table~\ref{tab:pacs}, our framework outperforms the state-of-the-art methods by an average of 0.86\% with a consistent margin.
Also, following DomainBed~\cite{Gulrajani21domainbed}, we report the domain generalization performance on various domain generalization benchmark datasets. 
In Table~\ref{tab:five}, our SDA shows a solid performance improvement across different DG datasets. 
Our method achieves state-of-the-art performance or the best SOTA competitor performance with a consistent margin on the out-of-distribution setting.
\input{Tables/3_Five}

\input{Tables/2_Ablations}

%% file: Figures/SD.tex
\begin{figure}[t]
    \centering
    \includegraphics[width=\columnwidth]{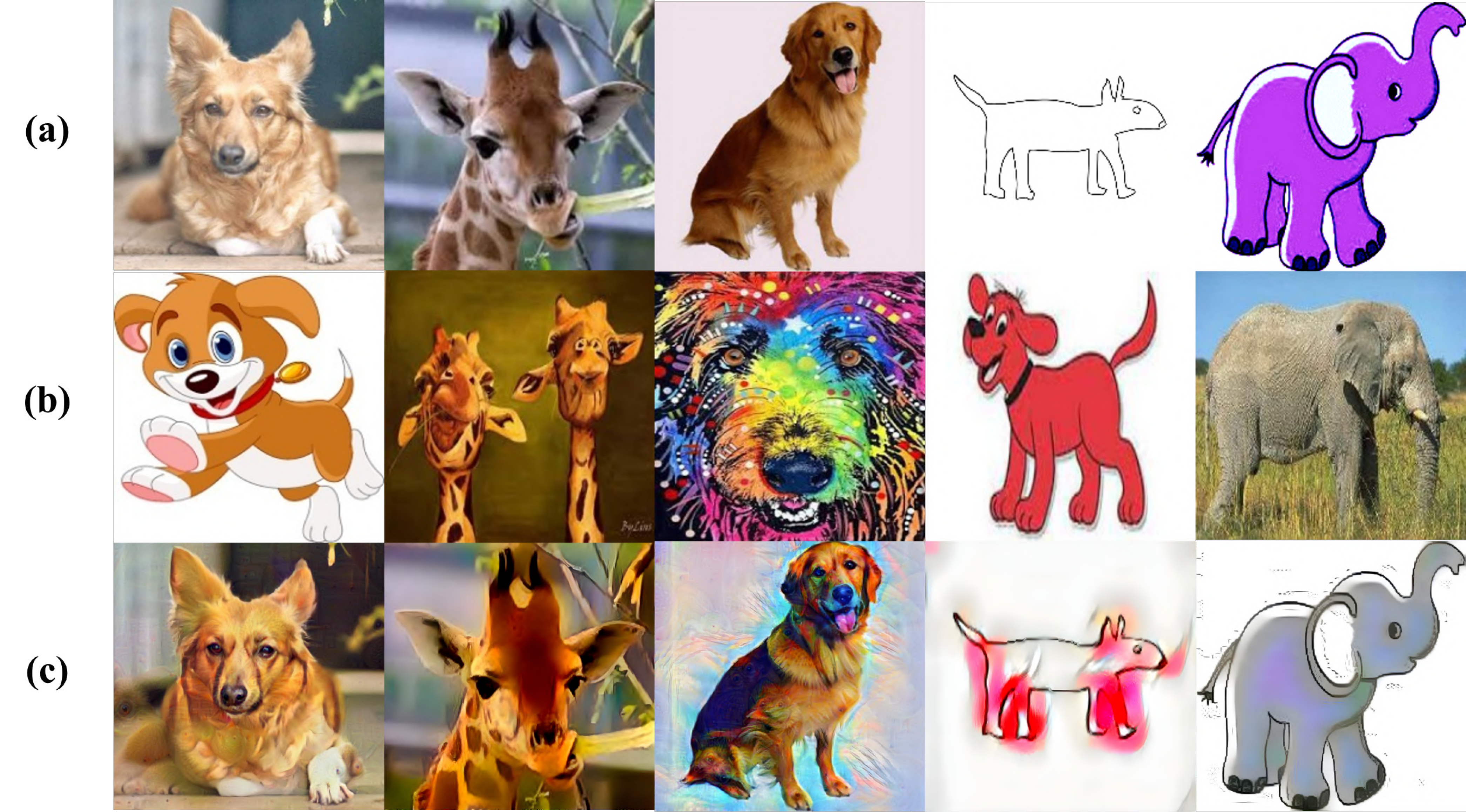}
    \caption{\textbf{Examples of Stlized Dreaming.} 
    Row (a) denotes the original image, row (b) denotes arbitrary target style image, and row (c) denotes the Stylized Dream image with its target style.   
    All the Stylized Dream examples are visualized by using VGGNet-16 feature extractor with 10 iterations and 0.09 of step size $\alpha$.
    Best viewed in color.
    }
    \label{fig:SD}
\end{figure}

%% file: Figures/shape.tex
\begin{figure}[t]
    \centering
    \includegraphics[width=0.8\columnwidth]{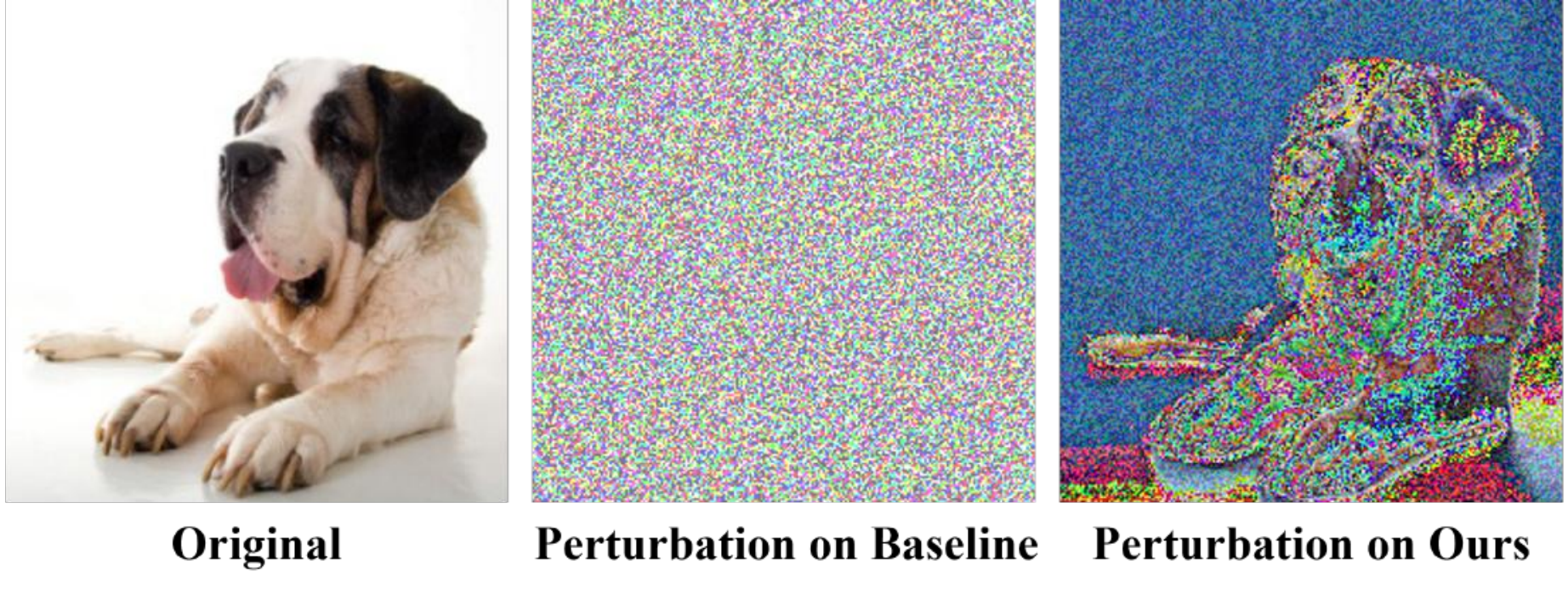}
    \caption{\textbf{Examination of Learned Representation of our model.}
    The left image denotes the original image for adversarial perturbation. 
    The Middle and the right image is adversarial perturbation on the baseline model and proposed method respectively.
    Each model of the adversarial attack is equal to rows (d) and (a) in Table~\ref{tab:ablation}.
    }
    \label{fig:adv}
\end{figure}

%% file: Tables/8_res_single.tex
\begin{table}[t!]
  \centering
  \small
  \caption{
  \textbf{Single-source domain generalization of ResNet-18 on PACS.  }
  We evaluate the domain generalization performance of the ResNet-18 model with ERM training on the single source domain (rows) and evaluate with other remaining target domains (columns).
  }
  \label{tab:res_single}
  \begin{adjustbox}{width=0.7\textwidth} 
  \begin{tabular}{ c  c c c c c  }
    \toprule
    \multirow{2}{*}{$\quad $ \textbf{ResNet-18} $\quad $} 
    & \multicolumn{4}{c}{{\textbf{Target Domain}}}  
    & \multirow{2}{*}{$\quad $ \textit{Average} $\quad $}\\ 
    \cmidrule(r){2-5} 
    & {P} &  {A} & {C} &  {S} &  \\ \midrule 
    
    \textit{Photo} & 
    - & $\ $ 63.87 $\ $  & $\ $ 33.02 $\ $  & $\ $ 31.76 $\ $ & \multicolumn{1}{c}{$\ $ 40.02 $\ $ } \\
    
    $\ $ \textit{Art Painting} $\ $ & 
    $\ $ 91.44 $\ $ & - & 62.66 & 66.42 & \multicolumn{1}{c}{69.82} \\
    
    \textit{Cartoon} & 
    79.46 & 61.28 & - & 71.67 & \multicolumn{1}{c}{70.59} \\
    
    \textit{Sketch} & 
    34.13 & 38.67 & 55.97 & - & \multicolumn{1}{c}{44.11} 
    \\
    \bottomrule
  \end{tabular}
  \end{adjustbox}
  \vspace{3pt}
\end{table}

%% file: Tables/1_Single.tex
\begin{table}[t!]
  \centering
  \small
  \caption{
  \textbf{Single-source Domain Generalization performance on PACS. }
  We report the performance of our framework with ResNet-18 backbone network on the single source domain (rows) and evaluate with other remaining target domains (columns).
  As a baseline, we compare with SelfReg~\cite{Kim21selfreg} in the same setting of the backbone network (compare left and right tables).
  }
  \label{tab:single}
  \begin{adjustbox}{width=1\textwidth} 
  \begin{tabular}{ c  c c c c c   c c c c c c }
    \toprule
    
    \multirow{2}{*}{\textbf{SelfReg~\cite{Kim21selfreg}}} & \multicolumn{5}{c}{{Target Domain}} &  
    \multirow{2}{*}{\textbf{Ours}} & \multicolumn{5}{c}{{Target Domain}} \\ 
    \cmidrule(r){2-6}  \cmidrule(r){8-12} 
    & {P} &  {A} & {C} &  {S} & \textit{Avg.} & 
    & {P} &  {A} & {C} &  {S} & \textit{Avg.} \\  \midrule 
    
    \textit{Photo} & 
    - & \textbf{67.72} & $\ $ 28.97 $\ $ & $\ $ 33.71 $\ $ & \multicolumn{1}{c|}{$\ $ 43.46 $\ $} &
    \textit{Photo} & 
    - & {$\ $ 67.24 $\ $} & \textbf{36.18} & \textbf{34.18}  & \textbf{45.87}  \\
    
    $\ $ \textit{Art Painting} $\ $ & 
    $\ $ {96.62} $\ $ & - & 65.22  & 55.94  & \multicolumn{1}{c|}{$\ $ 72.59 $\ $}  &
    $\ $ \textit{Art Painting} $\ $ & 
      \textbf{96.69}  & - & \textbf{66.68} & \textbf{56.98}  & \textbf{73.45} \\
    
    \textit{Cartoon} & 
    {87.53} & 72.09 & - & 70.06 & \multicolumn{1}{c|}{76.56} &
    \textit{Cartoon} & 
    \textbf{88.14}  & \textbf{73.78}  & - & \textbf{74.85} & \textbf{78.92} \\
    
    \textit{Sketch} & 
    {$\ $ 46.07 $\ $} & 37.17 & 54.03 & - & \multicolumn{1}{c|}{45.76} &
    \textit{Sketch} & 
    \textbf{52.87}  & \textbf{52.10}  & \textbf{63.05}  & - & \textbf{56.01}  \\ 
    \midrule
    
    \textit{Average} & 
    {76.74} & 58.99 & 49.41 & 53.24 & \multicolumn{1}{c|}{59.59} & 
    \textit{Average} & 
    \textbf{79.23}  & \textbf{64.37}  & \textbf{55.30}  & \textbf{55.33} 
    & \textbf{63.56} \\
    \bottomrule
  \end{tabular}
  \end{adjustbox}
  \vspace{3pt}
\end{table}

%% file: Tables/0_Main.tex
\begin{table}[ht!]
    \centering
    \small
    \caption{
    \textbf{Comparison with the state-of-the-art domain generalization methods and SDA on PACS dataset.} 
    All experiments are evaluated in out-of-distribution setting using ResNet-18 and ResNet-50 architectures. 
    The title of each column indicates the test target domain.
    We report RSC*~\cite{Huang20RSC} as reproducible performance. DSON*~\cite{Seo20DSON} denotes DSON model with domain-specific mixture-weights, \textit{i.e.}, model ensemble performance. 
    \textbf{Bold} denotes the best performance.}
    \label{tab:pacs}
    \begin{tabular}{l|c c c c|c} 
    \toprule
    \multicolumn{1}{l|}{\textbf{ Method}} & \multicolumn{1}{c}{$\ \ $Art Paint. $\ $} & \multicolumn{1}{c}{$\ \ $ Cartoon $\ \ $} & \multicolumn{1}{c}{$\ \ $ Photo $\ \ $} & \multicolumn{1}{c|}{$\quad $  Sketch  $\quad $}& {\textbf{ Average (\%) }}  \\  
    \hline 

    MetaReg~\cite{Balaji18metareg} 
    & 87.20  &  79.20  &  97.60 &  70.30 & 83.60 \\
    
    DANN~\cite{ganin2017dann}
    & 86.4 & 77.4 & 97.3 & 73.5 & 83.6 \\
    
    CDANN~\cite{li2018cdann}
    & 84.6 & 75.5 & 96.8 & 73.5 & 82.6 \\
    
    MTL~\cite{blanchard2021mtl} 
    & 87.5 & 77.1 & 96.4 & 77.3 & 84.6 \\
    
    VREx~\cite{krueger2021vrex} 
    & 86.0 & 79.1 & 96.9 & 77.7 & 84.9 \\
    
    MSAF~\cite{Dou19msaf} 
    & 82.89 & 80.49 & 95.01 & 72.29 & 82.67 \\
    
    RSC*~\cite{Huang20RSC} 
    & 81.38 & 80.14 & 93.72 & 82.31 & 84.38 \\
    
    ER~\cite{zhao20er} 
    & {87.51} & 79.31 & \textbf{98.25} & 76.30 & 85.34  \\
    
    CORAL~\cite{sun2016coral} 
    & 88.30 & 80.00 & 97.50 & 78.80 & 86.20  \\
    
    ARM~\cite{zhang2020arm} 
    & 86.80 & 76.80 & 97.40 & 79.30 & 85.10  \\
    
    Mixup~\cite{yan2020mixup} 
    & 86.10 & 78.90 & 97.60 & 75.80 & 84.60  \\
    
    pAdaIN~\cite{nuriel21padain} 
    & 85.82 & 81.06 & 97.17 & 77.37 & 85.36 \\
    
    IRM~\cite{arjovsky2019irm} 
    & 84.80 & 76.40 & 96.70 & 76.10 & 83.50 \\
    
    GroupDRO~\cite{sagawa2020groupdro} & 83.5 & 79.1 & 96.7 & 78.3 & 84.4 \\
    
    MLDG~\cite{li2018mldg} 
    & 85.5 & 80.1 & 97.4 & 76.6 & 84.9 \\
    
    ERM~\cite{Gulrajani21domainbed} 
    & 84.70 & 80.80 & 97.20 & 79.30 & 85.50 \\
    
    EISNet~\cite{wang20eisnet} 
    & 86.64 & 81.53 & 97.11 & 78.07 & 85.84 \\
    
    CORAL~\cite{sun2016coral}
    & 88.3 & 80.0 & 97.5 & 78.8 & 86.2 \\
    
    MMD~\cite{Gulrajani21domainbed}
    & 86.1 & 79.4 & 96.6 & 76.5 & 84.6 \\

    
    DSON*~\cite{Seo20DSON} 
    & 87.04 & 80.62 & 95.99 & {82.90} & 86.64  \\
    
    SelfReg~\cite{Kim21selfreg} 
    & 87.90 & 79.40 & 96.80 & 78.30 & 85.60  \\
    
    SagNet~\cite{Nam21sagnet} 
    & 87.40 & 80.70 & 97.10 & 80.00 & 86.30  \\
    
    SWAD~\cite{cha2021swad}
    & {89.3} & {83.4} & 97.3 & 82.5 & {88.1} \\ 
    \hline 
    
    
    
    

    
    \multirow{2}{*}{\textbf{Ours}$_\text{SDA}$}
    & \textbf{90.23} 
    & \textbf{84.88} 
    & {96.77}
    & \textbf{83.96} 
    & {\textbf{88.96}}  \\
    
    & ($\pm$0.52)
    & ($\pm$0.24)
    & ($\pm$0.93)
    & ($\pm$0.05)
    & ($\pm$0.33) \\
    
    
    \bottomrule
  \end{tabular}
  \vspace{3pt}
\end{table}

%% file: Tables/3_Five.tex
\begin{table*}[t!]
    \centering
    \caption{\textbf{Comparision with the state-of-the-art domain generalization methods and SD on public DG benchmark dataset.} Average out-of-domain accuracies on five domain generalization benchmarks in the DomainBed setting. 
    We highlight the best results for each dataset.
    \textbf{Bold} and $\dagger$ mean the best performance and the second best performance respectively.
    }
  \label{tab:five}
  \begin{adjustbox}{width=\textwidth}
    \begin{tabular}{l|c c c c c|c} 
    \toprule
    \multicolumn{1}{l|}{\textbf{Method}} & 
    
    \multicolumn{1}{c}{$\ $ PACS $\ $} & 
    \multicolumn{1}{c}{$\ $ VLCS $\ $} & 
    \multicolumn{1}{c}{$\ $ OfficeHome $\ $}& 
    \multicolumn{1}{c}{$\ $ TerraInc $\ $} & 
    \multicolumn{1}{c|}{$\ $  DomainNet  $\ $}
    
    & {\textbf{ Average }} \\  
    \hline
    
    
    MMD~\cite{akuzawa2019mmd} 
    & $\ $ 84.7 $\ $ & $\ $ 77.5 $\ $ & $\ $ 66.4 $\ $ & $\ $ 42.2 $\ $ & $\ $ 23.4 $\ $  
    & $\ $ 58.8 $\ $ \\ 
    
    Mixstyle~\cite{zhou21mix} 
    & 85.2 & 77.9 & 60.4 & 44.0 & 34.0 
    & 60.3  \\  
    
    GroupDRO~\cite{sagawa2020groupdro} 
    & 84.4 & 76.7 & 66.0 & 43.2 & 33.3 & 60.7  \\
    
    IRM~\cite{arjovsky2019irm} 
    & 83.5 & 78.6 & 64.3 & 47.6 & 33.9 & 61.6  \\
    
    ARM~\cite{zhang2020arm} 
    & 85.1 & 77.6 & 64.8 & 45.5 & 35.5 & 61.7  \\
    
    VREx~\cite{krueger2021vrex} 
    & 84.9 & 78.3 & 66.4 & 46.4 & 33.6 & 61.9  \\
    
    CDANN~\cite{li2018cdann} 
    & 82.6 & 77.5 & 65.7 & 45.8 & 38.3 & 62.0  \\ 
    
    DANN~\cite{ganin2017dann} 
    & 83.7 & 78.6 & 65.9 & 46.7 & 38.3 & 62.6  \\
    
    RSC~\cite{Huang20RSC} 
    & 85.2 & 77.1 & 65.5 & 46.6 & 38.9 & 62.7  \\ 
    
    MTL~\cite{blanchard2021mtl} 
    & 84.6 & 77.2 & 66.4 & 45.6 & 40.6 & 62.9  \\ 
    
    ERM~\cite{Gulrajani21domainbed} 
    & 85.5 & 77.5 & 66.5 & 46.1 & 40.9 & 63.3  \\
    
    Mixup~\cite{yan2020mixup} 
    & 84.6 & 77.4 & 68.1 & 47.9 & 39.2 & 63.4  \\
    
    MLDG~\cite{li2018mldg} 
    & 84.9 & 77.2 & 66.8 & 47.8 & 41.2 & 63.6  \\ 
    
    SagNet~\cite{Nam21sagnet} 
    & 86.3 & 77.8 & 68.1 & 48.6 & 40.3 & 64.2  \\ 
    
    CORAL~\cite{sun2016coral} 
    & 86.2 & {78.8} & 68.7 & 47.7 & 41.5 & 64.5  \\ 
    
    SelfReg~\cite{Kim21selfreg} 
    & 85.6 & 77.8 & 67.9 & 47.0 & 42.8 & 64.2  \\ 
    
    SWAD~\cite{cha2021swad} 
    & \textbf{88.1}${}^{\dagger}$ & \textbf{79.1} & \textbf{70.6} & \textbf{50.0}${}^{\dagger}$ & \textbf{46.5} & \textbf{66.9}${}^{\dagger}$  \\ 
    \hline

    \multirow{2}{*}{\textbf{Ours}}
    & \textbf{89.0} 
    & \textbf{78.8}${}^{\dagger}$
    & \textbf{70.4}${}^{\dagger}$
    & \textbf{51.1}
    & \textbf{46.4}${}^{\dagger}$ 
    & \textbf{67.2}
    \\
     
    & ($\pm$0.3)
    & ($\pm$0.1)
    & ($\pm$0.3)
    & ($\pm$1.2) 
    & ($\pm$0.7) 
    & ($\pm$1.1)
    \\
 
    \bottomrule
  \end{tabular}
  \end{adjustbox}
  \vspace{3pt}
\end{table*}

%% file: Tables/2_Ablations.tex
\begin{table*}[!ht]
  \centering
  \small
  \caption{\textbf{Ablation Study of SD augmentation on PACS dataset.} \textit{Abbr.}
  \textit{Cons. Reg.}: Consistency regularization between SD and clean image. 
  In row (c), we directly feed data augmentations Stylized Dream to cross-entropy loss. 
  }
  \label{tab:ablation}
  \begin{adjustbox}{width=\textwidth}
  \begin{tabular}{ c  c c c  c c c c  c  }
    \toprule
    \multirow{2}{*}{} & \multicolumn{3}{c}{\textbf{Components}} & \multicolumn{4}{c}{\textbf{Test Domain}} &
    \multirow{2}{*}{$\ $ \textit{Average}(\%)}
    \\
    \cmidrule{2-4} \cmidrule{5-8}
     & $\ $\textit{Stylized Dream} $\ $
     & $\ $\textit{Deep Dream} $\ $
     & $\ $\textit{Cons. Reg.} $\ $\hspace{1pt} 
     & {A} 
     & {C} 
     & {S} 
     & {P} 
     \\  \midrule
     
    (a)
    & \checkmark &  & \checkmark &      
    $\ $ \textbf{90.23} $\ $  
    & $\ $ \textbf{84.88} $\ $ 
    & $\ $ \textbf{83.96} $\ $ 
    & $\ $ {96.77} $\ $ 
    & $\ $ \textbf{88.96} $\ $ 
    \\ 
    
    (b)
    &   & \checkmark  & \checkmark
    & 88.18 
    & 79.86 
    & 78.34 
    & \textbf{97.49} 
    & 85.97 
    \\
    
    (c)
    & \checkmark  &    & 
    & 85.09 
    & 81.66 
    & 79.13 
    & 96.75 
    & 85.66 
    \\

    (d)
    &  &  &
    & 85.16 
     & 78.89 
     & 78.04 
     & 95.07 
     & 84.29 
     \\
    
    \bottomrule
  \end{tabular}
  \end{adjustbox}
  \vspace{3pt}
\end{table*}

    
    
    
    
    
    
    

%% file: 4_Experiments/4_3_Ablations.tex
\subsection{Ablation Studies}
\label{subsec:ablation}
In Table~\ref{tab:ablation}, we compare variants of our training strategy.
We show the effectiveness of each component in our method by, 
(a) Consistency training with SD (ours), 
(b) Consistency training with DeepDream,
(c) ERM training with SD, \textit{i.e.,} trained as a CE loss with both clean and SD images. 
(d) Baseline (ResNet50) ERM training without SD in the out-of-distribution setting.

\subsubsection{{Effect of \textit{Stylized Dream.}}} 

To verify the effectiveness of the proposed learning framework, we first train our model with only the supervised loss, utilizing the proposed \textit{Stylized Dream} as an augmentation.
In other words, we augment all the training images with Stylized Dreaming and directly feed the augmentations to the supervised loss. 
As shown in rows (c) and (d) in Table~\ref{tab:ablation}, we observe that our \textit{Stylized Dream} improves the domain generalization performance by an average of 1.37\%. 
It shows performance improvement across all the domains except \textit{art painting}, though it shows competitive performance.
Such improvement is noticeable in Cartoon domain where the performance gain is 2.76\%.

\subsubsection{{Effect of Consistency Regularization.} } 

Second, we leverage Stylized Dream with consistency regularization, instead of directly feeding it to the supervised loss. 
As discussed in Section~\ref{sec:intro} and ~\ref{sec:SDA_train}, learning the shape-based representation to be an important inductive bias gives more generalizability. 
In contrast to the training network only with the Stylized Dream and its supervised loss, the consistency regularization directly makes the model find the commonality between original images and Stylized Dream.
In the row (a) and row (c) of Table~\ref{tab:ablation}, the network trained with consistency regularization outperforms without consistency regularization, with 3.3\% of performance improvement. 
This implies that our consistency regularization with Stylized Dream encourages the network to integrate long-range spatial information, \textit{i.e.}, shape.

\subsubsection{{Effect of AdaIN in \textit{Stylized Dream}.}} 
We now experimentally show that SD without AdaIN, \textit{i.e.}, \textit{DeepDream}, is not enough to improve the performance of the domain generalization. 
According to Zhou et al.~\cite{zhou21mix}, a style of an image is closely related to the domain, so augmenting images by replacing styles while preserving contents results in better domain generalizability.  
In Table~\ref{tab:ablation} row (a) and (b), our method outperforms training with \textit{DeepDream} in the overall domain by 2.99\% on average.
This is a clear indication that the Stylized Dream can actually reduce local texture cues, while Deep Dream isn't enough. 
In other words, this implies that feature enhancement with AdaIN prevents the exploitation of style-based representation, which leads to improved domain generalization performance.
\input{Tables/7_divergence}

\subsubsection{{Consistency Regularization Metrics.}} 
In Equation~\ref{eq:consistency_reg}, there are various options for the divergence metric. 
In Table~\ref{tab:divergence}, we compared different divergence metrics on our training framework. 
First, we use a mean-squared error between the output distribution of the original image and that of Stylized Dream. 
The second metric and third metric are KL divergence and JS divergence between output distribution of the original image and that of Stylized Dream respectively. 
Over the experiments, ResNet-50 trained with KL divergence metric achieves the best performance than the same model trained with mean-squared error and JS divergence.
Therefore, we utilized the KL divergence metric in all other experiments of the paper.
\input{Tables/5_Baseline}

\subsubsection{{Evaluation of our method on various backbone networks.}} 
Lastly, we show the scalability and robustness of our method by reporting experiments of our method on various backbone networks; VGGNet-16, ResNet-18, and ResNet-50. 
In Table~\ref{tab:baseline}, our method all outperforms vanilla ERM training regardless of the model selection.  
Our method consistently improves the performance of domain generalization across the backbone networks, which implies the scalability of our method. 
Also, a large margin across the domains supports the robustness of our method.

%% file: Tables/7_divergence.tex
\begin{table*}[t!]
  \centering
  \small
  \caption{\textbf{Consistency Training with various regularization metric.} 
  We report the experiments of our consistency regularization with 3 divergence metrics: Mean-Squared Error, Jensen-Shannon Divergence, and Kullback–Leibler divergence. `-' denotes ERM for the comparison.
  }
  \label{tab:divergence}
  \begin{adjustbox}{width=\textwidth}
  \begin{tabular}{ c  c   c c c c  c  }

    \toprule
    \multirow{2}{*}{$\quad$  \textbf{Model}  $\quad$ } 
    & \multirow{2}{*}{$\quad$ \textbf{Divergence Metric} $\quad$}
    & \multicolumn{4}{c}{\textbf{Test Domain}} 
    & \multirow{2}{*}{$\quad$ \textit{Average}(\%) $\quad$}
    \\
    \cmidrule{3-6} 
     &
     & {A} 
     & {C} 
     & {S} 
     & {P} \hspace{1pt} 
     \\  \midrule
    
    \multirow{4}{*}{$\ $ ResNet-50} 
    & -
    & $\ $ 85.16 $\ $ 
    & $\ $ 78.89 $\ $ 
    & $\ $ 78.04 $\ $ 
    & {95.07} 
    & $\ $ 84.29 $\ $ 
    \\

    & Mean-Squared Error 
    & $\ $ 87.70 $\ $ 
    & $\ $ 82.76 $\ $ 
    & $\ $ 81.90 $\ $ 
    & \textbf{97.49} 
    & $\ $ 87.46 $\ $ 
    \\
     
    &  JS-Divergence &
    {89.48}  & 
    {82.89}  & 
    {81.01}  & 
    {97.25}  & 
    {87.66}  \\ 

    &  KL-Divergence &
    \textbf{90.23}  & 
    \textbf{84.88}  & 
    \textbf{83.96}  & 
    $\ $ {96.77} $\ $  & 
    \textbf{88.96}  \\ 
    
    \bottomrule
  \end{tabular}
  \end{adjustbox}
  \vspace{3pt}
\end{table*}

%% file: Tables/5_Baseline.tex
\begin{table*}[!ht]
  \centering
  \small
  \caption{\textbf{Our framework in various baseline Models.} Average out-of-domain accuracies with different baseline models; VGGNet-16, ResNet-18, and ResNet-50. We report each performance of backbone networks compared with ERM training.
  }
  \label{tab:baseline}
  \begin{adjustbox}{width=\textwidth}
  \begin{tabular}{ c  c   c c c c  c  }
    \toprule
    \multirow{2}{*}{$\quad$  \textbf{Model}  $\quad$ } 
    & \multirow{2}{*}{$\quad$ \textbf{Method} $\quad$}     
    & \multicolumn{4}{c}{\textbf{Test Domain}} 
    & \multirow{2}{*}{$\quad$ \textit{Average}(\%) $\quad$}
    \\
    \cmidrule{3-6} 
     & 
     & {A} 
     & {C} 
     & {S} 
     & {P} \hspace{1pt} 
     \\  \midrule
     
    \multirow{2}{*}{$\quad $  VGGNet-16 $\quad $ }
    &   ERM 
    & $\ $ 81.03 $\ $
    & $\ $ 77.82 $\ $
    & $\ $ 76.40 $\ $
    & $\ $ 94.17 $\ $
    & $\ $ 82.35 $\ $
    \\
    
    & $\quad $ Ours $\quad $ 
    & \textbf{86.08} 
    & \textbf{81.83} 
    & \textbf{81.22} 
    & \textbf{96.65} 
    & \textbf{86.52} 
    \\ \midrule
    
    \multirow{2}{*}{ResNet-18}
    &  ERM
    & 79.89 
    & 75.61 
    & 73.33 
    & 95.66 
    & 81.12 
    \\

    &  Ours
    & \textbf{84.47}
    & \textbf{78.92}
    & \textbf{77.98}
    & \textbf{95.75}
    & \textbf{84.28}  \\ \midrule
    
    \multirow{2}{*}{ResNet-50}
    & ERM
    & 85.16 
    & 78.89 
    & 78.04 
    & 95.07 
    & 84.29 
    \\

    & Ours &
    \textbf{90.23}  & 
    \textbf{84.88}  & 
    \textbf{83.96}  & 
    \textbf{96.77}  & 
    \textbf{88.96}  \\ 
    
    \bottomrule
  \end{tabular}
  \end{adjustbox}
  \vspace{3pt}
\end{table*}

%% file: Content/5_Conclusions.tex
\section{Conclusion}
\label{sec:conclusion}
In this paper, we observe that the texture bias negatively affects not only in-domain generalization but also out-of-distribution generalization, \textit{i.e.}, Domain Generalization. 
From the observation, we propose \textit{Stylized Dream}, a novel optimization-based data augmentation method for domain generalization by alleviating the texture bias of DNNs.
\textit{Stylized Dream} intensifies the activation of an input image that is aligned with the style of the target image by AdaIN. 
This replaces the style of the original image while preserving the content.
We then adopt a consistent regularization between original images and Stylized Dream to make the model less texture biased. 
Experiments on various benchmark datasets show that our framework achieves state-of-the-art or competitive performance.

%% file: Content/6_Supplementary.tex
\newpage
\appendix
\renewcommand{\thesection}{\Alph{section}}

\section*{Appendix}
In this appendix, we first present additional results for the hyperparameter selection of our framework.
Second, we visualize Stylized Dream with the deep residual network that was not included in the main paper. 
Lastly, we provide the detailed experimental settings and PyTorch-like pseudo-code of \textit{Stylized Dream}. 
\section{Parameter Search}
\subsection{Step Size $\alpha$}
In Table~\ref{table:alpha}, we report the results of our framework on PACS datasets with various step size $\alpha$. 
All the experiments are conducted with the ResNet-50 backbone network. 
Overall, our framework is robust to a choice of step size $\alpha$.
\input{6_Supple/Tab_Alpha}

\subsection{Temperature parameter  $\tau$}
In Table~\ref{table:tem}, we report the results of our framework on the PACS dataset with various temperature parameters $\tau$ for three divergence metrics. 
We conduct all experiments with step size $\alpha=0.3$ and ResNet-50 as the backbone network.
As shown in Table~\ref{table:tem}, Photo and Art Painting domains are sensitive to the choice of the temperature parameter. 
With lower $\tau$, out-of-distribution accuracy on Photo and Art Painting shows significant degradation.
On the other hand, Cartoon and Sketch domains are robust to the choice of the temperature parameter.  
Interestingly, unlike other domains, the performance on the Sketch domain improves as the value of the temperature gets smaller, \textit{e.g.},  $\tau \rightarrow 1$.\input{6_Supple/Tab_tem}

\section{Visualization of Stylized Dream with ResNet}
We provide visualization of our \textit{Stylized Dream} across the different domains and classes in PACS dataset with a deep residual network.
The visualization of our model is shown in Fig~\ref{fig:SD2}.
To compare VGGNet and ResNet, Stylized Dream with ResNet-18 is visualized with the same examples as the main paper.
The figure has four rows showing the original images, arbitrary style images, its \textit{Stylized Dreams} with ResNet-18 backbone, and its \textit{Stylized Dreams} with VGGNet-16 backbone respectively. 
All the Stylized Dream examples are visualized with step size $\alpha=0.09 $ and $10$ iterations. 
As discussed in the literature~\cite{Gatys16transfer}, the features of deep residual networks are less effective for style transfer or stylization.
%

\input{6_Supple/code}

%% file: 6_Supple/Tab_Alpha.tex
\begin{table}[ht]
\renewcommand\thetable{8}
    \centering
    \caption{
    \textbf{Performance Evaluation with different choice of $\alpha$.} 
    All experiments are evaluated in an out-of-distribution setting using ResNet-50. 
    Temperature parameter $\tau$ for consistency regularization is fixed to 10. 
    The title of each column indicates the test target domain.
    \textit{Abbr.} A: Art Painting, C: Cartoon, S: Sketch, P: Photo.  
    }
    \label{tab:alpha}
    \begin{adjustbox}{width=\columnwidth}
    \begin{tabular}{c c  c c c c  c}
        \toprule
        
        \multirow{2}{*}{Divergence} & 
        \multirow{2}{*}{$\quad $ coefficient $\quad $} & 
        \multicolumn{4}{c}{Target Domain} & 
        \multirow{2}{*}{$\quad $ \textit{Average} \% $\quad $} \\
        \cmidrule{3-6}
        
        & $\alpha$
        & A 
        & C 
        & S 
        & P 
        & 
        \\
        \midrule
        
        \multicolumn{1}{c|}{}
        & \multicolumn{1}{c|}{0.01}
        & $\ $ {88.42} $\ $
        & $\ $ {83.43} $\ $ 		
        & $\ $ {82.05} $\ $
        & $\ $ {96.42} $\ $
        & {87.58} \\ 
        
        \multicolumn{1}{c|}{}
        & \multicolumn{1}{c|}{0.03}
        & {88.13}
        & {83.47}
        & {82.62}
        & {96.24}
        & {87.62} \\ 
        
        \multicolumn{1}{c|}{}
        & \multicolumn{1}{c|}{0.09}
        & {89.64} 
        & {83.59} 
        & {83.47}
        & {96.30} 
        & {88.25} \\ 
        
        \multicolumn{1}{c|}{$\ \ $  KL-Divergence  $\ $ }
        & \multicolumn{1}{c|}{0.15}
        & {89.99} 
        & {84.09}
        & \textbf{83.96}
        & {96.70}
        & {88.61} \\ 
        
        \multicolumn{1}{c|}{}
        & \multicolumn{1}{c|}{0.3}
        & \textbf{90.23}  
        & \textbf{84.88}  
        & \textbf{83.96}  
        & \textbf{96.77}
        & \textbf{88.96} 
          \\ 
        
        \multicolumn{1}{c|}{}
        & \multicolumn{1}{c|}{0.6}
        & {89.94} 
        & {84.69}
        & {81.59}
        & {96.06}
        & {88.07} \\ 
        
        \multicolumn{1}{c|}{}
        & \multicolumn{1}{c|}{0.9}
        & {89.55}
        & {84.17}
        & {79.43}
        & {95.58}
        & {87.21} \\ \midrule
        
        \multicolumn{1}{c|}{}
        & \multicolumn{1}{c|}{0.01}
        & {87.25} 
        & {83.36}
        & {80.85}
        & {96.18}
        & {86.91} \\ 
        
        \multicolumn{1}{c|}{}
        & \multicolumn{1}{c|}{0.03}
        & {88.81}
        & {84.39}
        & {81.06}
        & {96.48}
        & {87.69} \\ 
        
        \multicolumn{1}{c|}{}
        & \multicolumn{1}{c|}{0.09} 
        & {89.40} 
        & {83.89} 
        & {81.78} 
        & {96.65}
        & {87.93} \\ 
        
        \multicolumn{1}{c|}{$\ \ $  JS-Divergence  $\ $ }
        & \multicolumn{1}{c|}{0.15}
        & \textbf{90.23} 
        & {84.05}
        & \textbf{82.45}
        & {96.71}
        & \textbf{88.36} \\ 
        
        \multicolumn{1}{c|}{}
        & \multicolumn{1}{c|}{0.3}
        & {89.84}       
        & \textbf{84.09}   
        & {80.61}  
        & \textbf{97.42} 
        & {87.49}  \\ 
        
        \multicolumn{1}{c|}{}
        & \multicolumn{1}{c|}{0.6} 
        & {89.06}
        & {82.81}
        & {81.19}
        & {96.48}
        & {87.39} \\ 
        
        \multicolumn{1}{c|}{}
        & \multicolumn{1}{c|}{0.9} 
        & {89.26}
        & {82.24}
        & {80.26}
        & {96.95}
        & {87.18} \\ \midrule

        \multicolumn{1}{c|}{}
        & \multicolumn{1}{c|}{0.01} 
        & {88.67}
        & {83.75}
        & {80.57}
        & {96.42}
        & {87.35} \\ 
        
        \multicolumn{1}{c|}{}
        & \multicolumn{1}{c|}{0.03} 
        & {89.16}
        & {83.79}
        & {80.47}
        & {96.24}
        & {87.42} \\ 
        
        \multicolumn{1}{c|}{}
        & \multicolumn{1}{c|}{0.09} 
        & {89.16} 
        & {84.28} 
        & {81.08}
        & {96.48} 
        & {87.75} \\ 
        
        \multicolumn{1}{c|}{$\ \ $  Mean-Squared Error  $\ $ }
        & \multicolumn{1}{c|}{0.15} 
        & {89.89}
        & {83.53}
        & {81.40}
        & {96.30}
        & {87.78} \\ 
        
        \multicolumn{1}{c|}{}
        & \multicolumn{1}{c|}{0.3}
        & {88.69}       
        & {82.79}  
        & {81.9}  
        & \textbf{97.54} 
        & {87.73}  \\ 
        
        \multicolumn{1}{c|}{}
        & \multicolumn{1}{c|}{0.6}
        & \textbf{90.38}
        & \textbf{84.39}
        & \textbf{82.29}
        & {96.24}
        & \textbf{88.33} \\
        
        \multicolumn{1}{c|}{}
        & \multicolumn{1}{c|}{0.9}
        & {89.16} 	
        & {84.04}
        & {80.98}
        & {96.06}
        & {87.56} \\ 
        
        \bottomrule
    \end{tabular}
    \end{adjustbox}
    \label{table:alpha}
\end{table}

%% file: 6_Supple/Tab_tem.tex
\renewcommand\thefigure{3}
\begin{figure}[!t]
    \centering
    \caption{\textbf{Examples of Stlized Dreaming.} 
    (a) Original images (b) Arbitrary target style images (c) Stylized Dream images generated with ResNet-18, and (d) Stylized Dream images generated with VGGNet-16.   
    All the Stylized Dream examples are generated with step size $\alpha=0.09$ and 10 iterations .
    Best viewed in color.
    }
    \vspace{5pt}
    \includegraphics[width=\columnwidth]{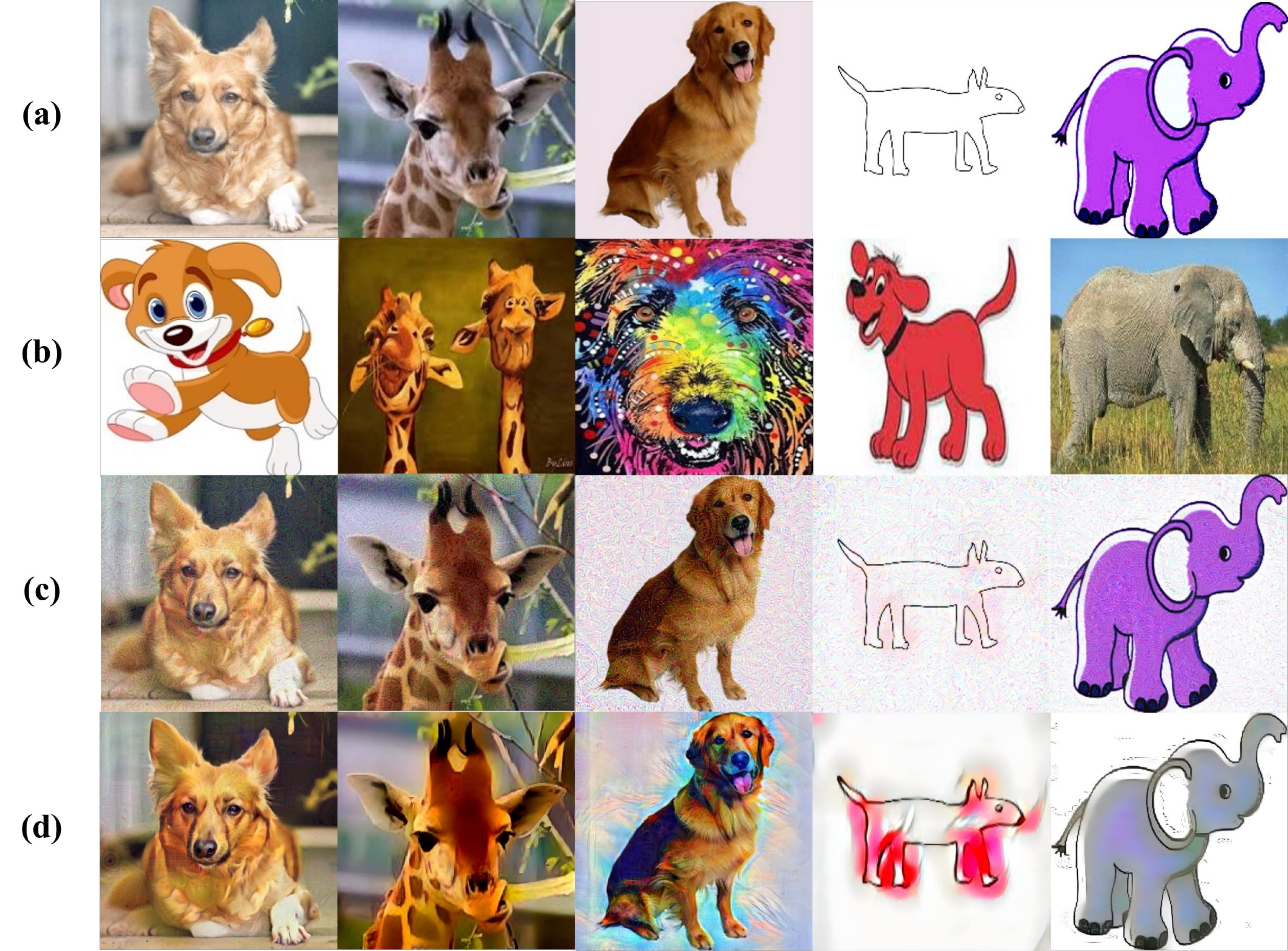}
    \label{fig:SD2}
\end{figure}

\begin{table}[!t]
\renewcommand\thetable{9}
    \centering
    \caption{
    \textbf{Performance Evaluation with different choice of temperature $\tau$.} 
    All experiments are evaluated in an out-of-distribution setting using ResNet-50. 
    Step size $\alpha$ of Stylized Dream is fixed to 0.03. 
    The column names indicate the test target domains.
    \textit{Abbr.} A: Art Painting, C: Cartoon, S: Sketch, P: Photo.  
    }
    \label{tab:tau}
    \begin{adjustbox}{width=\columnwidth}
    \begin{tabular}{c  c  c c c c  c}
        \toprule
        
        \multirow{2}{*}{$\quad $ Divergence $\quad $} &
        {$\quad $ Temperature $\quad \ $} & 
        \multicolumn{4}{c}{\textbf{Target Domain}} & 
        \multirow{2}{*}{$\quad $ \textit{Average} \% $\quad $} \\
        \cmidrule{3-6}
        
        & $\tau$ 
        & A 
        & C 
        & S 
        & P 
        & 
        \\
        \midrule
        
        \multicolumn{1}{c|}{ }
        & \multicolumn{1}{c|}{1}
        & {76.22} 			
        & {83.6}
        & \textbf{85.2}
        & {87.79}
        & {83.20} \\ 
        
        \multicolumn{1}{c|}{ }
        & \multicolumn{1}{c|}{3}
        & {87.59} 				
        & {84.00}
        & {84.26}
        & {94.68}
        & {87.63} \\

        \multicolumn{1}{c|}{$\ \ $  KL-Divergence  $\ $ }
        & \multicolumn{1}{c|}{5}
        & $\ $ {88.77} $\ $				
        & $\ $ {83.96} $\ $
        & $\ $ {84.42} $\ $
        & $\ $ {95.64} $\ $
        & {88.19} \\ 
        
        \multicolumn{1}{c|}{ }
        & \multicolumn{1}{c|}{10}
        & \textbf{90.23} 				
        & \textbf{84.88}
        & {83.96}
        & \textbf{96.77}
        & \textbf{88.96} \\ 
        
        \multicolumn{1}{c|}{ }
        & \multicolumn{1}{c|}{20}
        & {89.69} 				
        & {84.71}
        & {82.70}
        & {96.65}
        & {88.44} \\ \midrule
        
        \multicolumn{1}{c|}{ }
        & \multicolumn{1}{c|}{1}
        & {78.41} 				
        & {83.36}
        & \textbf{82.71}
        & {89.29}
        & {83.44} \\ 
        
        \multicolumn{1}{c|}{ }
        & \multicolumn{1}{c|}{3}
        & {87.35} 				
        & {83.39}
        & {81.49}
        & {94.8}
        & {86.76} \\

        \multicolumn{1}{c|}{$\ \ $  JS-Divergence  $\ $ }
        & \multicolumn{1}{c|}{5}
        & {88.81} 				
        & {83.88}
        & {81.54}
        & {95.52}
        & {87.44} \\ 
        
        \multicolumn{1}{c|}{ }
        & \multicolumn{1}{c|}{10}
        & \textbf{89.84} 				
        & \textbf{84.09}
        & {80.61}
        & \textbf{97.42}
        & \textbf{87.49} \\ 
        
        \multicolumn{1}{c|}{ }
        & \multicolumn{1}{c|}{20} 				
        & {89.25}
        & {83.45}
        & {79.83}
        & {96.48}
        & {87.26} \\ \midrule
        
        \multicolumn{1}{c|}{ }
        & \multicolumn{1}{c|}{1}
        & {83.69} 				
        & {83.83}
        & \textbf{83.6}
        & {93.84}
        & {86.24} \\ 
        
        \multicolumn{1}{c|}{ }
        & \multicolumn{1}{c|}{3}
        & {89.3} 				
        & {83.88}
        & {82.93}
        & {95.64}
        & {87.94} \\

        \multicolumn{1}{c|}{$\ \ $  Mean-Squared Error  $\ $ }
        & \multicolumn{1}{c|}{5}
        & \textbf{89.5} 				
        & \textbf{84.13}
        & {82.43}
        & {96.18}
        & \textbf{88.05} \\ 
        
        \multicolumn{1}{c|}{ }
        & \multicolumn{1}{c|}{10}
        & {88.69} 				
        & {82.79}
        & {81.9}
        & \textbf{97.54}
        & {87.73} \\ 
        
        \multicolumn{1}{c|}{ }
        & \multicolumn{1}{c|}{20} 				
        & {88.55}
        & {82.73}
        & {81.31}
        & {96.71}
        & {87.33} \\ 
        
        \bottomrule
    \end{tabular}
    \end{adjustbox}
    \label{table:tem}
\end{table}

%% file: 6_Supple/code.tex
\section{Experimental Settings and Pseudo Code}
\subsection{Licences}
\begin{itemize}
    \item We use Pytorch as our base framework.\footnote{Copyright (c) 2016-Facebook, Inc            (Adam Paszke). Licensed under the BSD-style License.} \vspace{3pt}
    \item We modify the codes from SelfReg: Self-supervised Contrastive Regularization for Domain Generalization: {https://github.com/dnap512/SelfReg}\footnote{Copyright (c) 2021 dnap512. Licensed under the MIT License.}. \vspace{3pt}
    \item We also modify the codes from DomainBed: 
    
    \url{https://github.com/facebookresearch/DomainBed}.\footnote{Copyright (c) Facebook, Inc., Licensed under the MIT license} \vspace{3pt}
    
    \item We modify the following repository to run on multiple GPUs:
    
    \url{https://github.com/facebookresearch/simsiam}\footnote{Copyright (c) Facebook, Inc., Licensed under the CC-BY-NC 4.0 License.}.
    
    \item All our experiments are conducted by NVIDIA RTX3090.
\end{itemize}
\input{6_Supple/code_AdaIN}

\subsection{Pseudo Code of Stylized Dream}
The Pytorch-like code for Adaptive Instance Normalization (AdaIN) and \textit{Stylized Dream} (SD) are in Algorithm~\ref{alg:adain} and Algorithm~\ref{alg:SDA}, respectively.
\input{6_Supple/code_SD}

%% file: 6_Supple/code_AdaIN.tex
\renewcommand\thealgorithm{3}
\begin{algorithm}[t]
\caption{PyTorch-like code for AdaIN}
\label{alg:adain}
\definecolor{codeblue}{rgb}{0.25,0.5,0.5}
\definecolor{codekw}{rgb}{0.85, 0.18, 0.50}

\begin{verbatim}
    def calc_mean_std(feat):
        N, C, H, W = feat.size()
        feat_mean = feat.view(N, C, -1).mean(dim=2)
        feat_var = feat.view(N, C, -1).var(dim=2)
        feat_std = feat_var.sqrt()
        return feat_mean, feat_std
        
    def AdaIN(con_feat, style_feat):
        """
        Args:
            cont_feat: input feature map
            style_feat: feature map to extract style
        Returns:
            AdaIN output
        """
        # calculate mean and std of feature maps
        con_mean, con_std = calc_mean_std(content_feat)
        style_mean, style_std = calc_mean_std(style_feat)
        
        # AdaIN
        normalized_feat = (con_feat - con_mean) / con_std
        return normalized_feat * style_std + style_mean
    end.
\end{verbatim}
\end{algorithm}

%% file: 6_Supple/code_SD.tex
\renewcommand\thealgorithm{4}
\begin{algorithm}[t]
\caption{PyTorch-like code for \textbf{\textit{Stylized Dream}}}
\label{alg:SDA}
\definecolor{codeblue}{rgb}{0.25,0.5,0.5}
\definecolor{codekw}{rgb}{0.85, 0.18, 0.50}

\begin{verbatim}
    def StylizedDream(model, image, target, noise, iterations, alpha):
        """
        Args:
            model: feature extractor
            image: an image that want to Stylize
            target: a target image for extracting style
            noise: noise boundary for an image
            iterations: number of iterations
            alpha: step size for Stylized Dream
        Returns:
            noise: image augmented with Stylized Dream
        """
        # imagenet RGB mean and std
        mean = array([0.485, 0.456, 0.406])
        std = array([0.229, 0.224, 0.225]) 
        # lower and upper bound for the image
        l_bound = -mean / std
        u_bound = (1 - mean) / std
        
        # arbitrary noise drawn from uniform
        # shape: same with image (3,224,224)
        n = uniform(-noise_bound, .noise_bound) 
        noise = (image+n).detach() 
        
        for i in range(iterations):
            noise.requires_grad = True
            model.zero_grad()
            
            # Perform AdaIN with two images
            c5_img = model.extract_features(noise)
            c5_tar = model.extract_features(target)
            c5_adain = AdaIN(c5_img, c5_tar.detach())

            # L2 norm of the output feature map
            loss = c5_adain.norm() 
            loss.backward()
            grad = noise.grad.data
            grad = (grad - grad.mean()) / grad.std() 
            
            # Update the image with the gradient
            SD = noise + lr * grad 
            # clip the image
            SD = max(min(SD, u_bound), l_bound) 
            SD = SD.detach()
            
        return SD
    end.
\end{verbatim}
\end{algorithm}